\pdfoutput=1

\documentclass[11pt]{article}

\usepackage[final]{acl}

\usepackage{times}
\usepackage{latexsym}

\usepackage{tcolorbox}
\tcbuselibrary{skins}
\usepackage{scrextend}
\usepackage[T1]{fontenc}

\usepackage[utf8]{inputenc}

\usepackage{microtype}

\usepackage{inconsolata}

\usepackage{graphicx}
\usepackage{booktabs}
\usepackage{amsmath}
\usepackage{amssymb}
\usepackage{multirow}
\usepackage{algorithmic}
\usepackage{algorithm}
\usepackage{listings}
\usepackage{enumitem}
\usepackage[table,xcdraw]{xcolor}
\usepackage{colortbl}

\definecolor{codegreen}{rgb}{0,0.6,0}
\definecolor{codegray}{rgb}{0.5,0.5,0.5}
\definecolor{codepurple}{rgb}{0.58,0,0.82}
\definecolor{backcolour}{rgb}{0.95,0.95,0.92}

\lstdefinestyle{mystyle}{
    backgroundcolor=\color{backcolour},   
    commentstyle=\color{codegreen},
    keywordstyle=\color{magenta},
    numberstyle=\tiny\color{codegray},
    stringstyle=\color{codepurple},
    basicstyle=\ttfamily\footnotesize,
    breakatwhitespace=false,         
    breaklines=true,                 
    captionpos=b,                    
    keepspaces=true,                 
    numbers=left,                    
    numbersep=5pt,                  
    showspaces=false,                
    showstringspaces=false,
    showtabs=false,                  
    tabsize=2
}

\usepackage[colorinlistoftodos,prependcaption]{todonotes}

\title{Identifying \& Interactively Refining Ambiguous \\ User Goals for Data Visualization Code Generation}

\author{
    Mert İnan\textsuperscript{1},
    Anthony Sicilia\textsuperscript{1}\footnotemark, 
    Alex Xie\textsuperscript{2}, \\
    \textbf{Saujas Vaduguru\textsuperscript{2}},
    \textbf{Daniel Fried\textsuperscript{2}},
    \textbf{Malihe Alikhani\textsuperscript{1}} \\
    \textsuperscript{1} Northeastern University,  Boston, MA, USA \\ 
    \textsuperscript{2} Carnegie Mellon University, Pittsburgh, PA, USA \\
    \texttt{\{inan.m, alikhani.m\}@northeastern.edu}
}

\begin{document}
\maketitle

\let\thefootnote\relax\footnotetext{* This author is currently at West Virginia University, but completed this work at Northeastern University.}
\begin{abstract}
Establishing shared goals is a fundamental step in human-AI communication. However, ambiguities can lead to outputs that seem correct but fail to reflect the speaker’s intent. In this paper, we explore this issue with a focus on the data visualization domain, where ambiguities in natural language impact the generation of code that visualizes data. The availability of multiple views on the contextual (\textit{e.g.} the intended plot and the code rendering the plot) allows for a unique and comprehensive analysis of diverse ambiguity types. We develop a taxonomy of types of ambiguity that arise in this task and propose metrics to quantify them. Using Matplotlib problems from the DS-1000 dataset, we demonstrate that our ambiguity metrics better correlate with human annotations than uncertainty baselines. Our work also explores how multi-turn dialogue can reduce ambiguity, and therefore, improve code accuracy by better matching user goals. We evaluate three pragmatic models to inform our dialogue strategies: Gricean Cooperativity, Discourse Representation Theory, and Questions under Discussion. A simulated user study reveals how pragmatic dialogues reduce ambiguity and enhance code accuracy, highlighting the value of multi-turn exchanges in code generation.
\end{abstract}

\begin{figure}[!ht]
    \centering
    \includegraphics[width=\columnwidth]{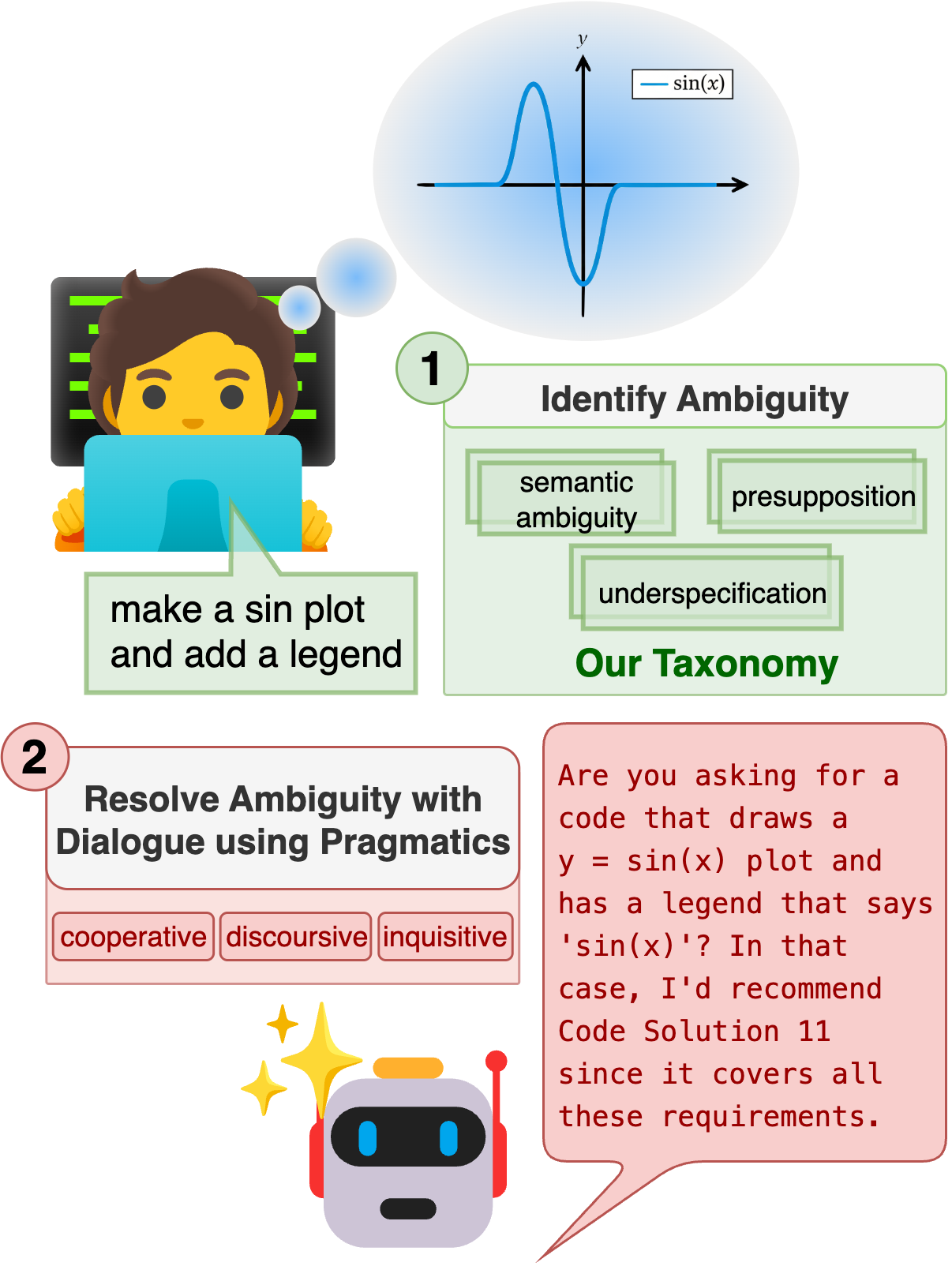}
    \caption{This figure summarizes the contributions of this paper. We formalize and identify ambiguity in data visualization code, then use pragmatics-inspired dialogue strategies to interactively resolve ambiguities in user intents. To this end, we present a multimodal taxonomy of ambiguity categories, and new metrics informed by this taxonomy.}
    \label{fig:main_fig}
    \vspace{-5pt}
\end{figure}






\section{Introduction}
\label{sec:intro}
In human-human interactions, ambiguity resolution has been explored through various well-established frameworks in linguistic pragmatics, such as Rational Speech Act (RSA)\cite{frank-goodman-2012-predicting}, Discourse Representation Theory (DRT) \cite{lascarides2007segmented, kamp2010discourse}, and Questions Under Discussion (QUD) \cite{roberts:2012:information}. Further, as \citet{Clark1996May} argues, successful interactions depend on establishing common ground by iterative alignment of assumptions and resolving ambiguities on the way to common ground. On the other hand, human-AI collaboration still often fails to provide a human-like pair-programming experience \cite{Williams, sarkar2022like}, and we claim that a key challenge here is ambiguity resolution—as users’ natural language descriptions of intent often map to multiple valid code implementations, requiring iterative clarifications. In this paper, we propose identifying (\S\ref{sec:methods}) and then resolving ambiguities (\S\ref{sec:dialogue}) in a dyadic setting of human-LLM interactions using pragmatics-inspired, persona-based prompting.


To achieve this goal, we frame the natural language to code problem as a two-player cooperative dialogue. A \textbf{director} (typically the user) specifies their intent in natural language and a \textbf{coder} (typically an automated coding assistant) generates code with the functionality the director had in mind. A goal for this pair-programming setting is to have a coder agent that can interact with the director agent to resolve ambiguity and generate code. While, in principle, the coder's uncertainty in this task can come from many sources, focus is typically placed on the inherent model limitations of the coder agent caused by insufficient knowledge or training. In contrast, this work focuses on the uncertainty of the coder about the user's goals---\textit{i.e.}, the \textit{ambiguity} of director's requests. We focus on how the director can resolve the coder's uncertainty through clear communication about intended goals, and how coders can clarify goals through conversation.

Based on our definition of ambiguity, we hypothesize that ambiguity reduction should lead to improved code accuracy. To study this, we propose a taxonomy that more precisely codifies what is meant by ambiguity. Our taxonomy includes categories of ambiguity, and examples specific to the plotting domain --- where the presence of multi-modal contexts allows us to explore more diverse sources of natural language ambiguity (\textit{e.g.}. We propose a number of metrics based on our taxonomy that allow us to dynamically (and automatically) quantify natural language ambiguity in text-to-code problems. We use our definition of ambiguity to study how multi-step communication with the director can serve to reduce ambiguity. Guided by theories of pragmatics we simulate dialogue between two machine agents. This shows how (pragmatic) dialogues can help coders resolve ambiguity, and ultimately, improve task success.

The contributions of this paper are as follows:
\begin{enumerate}[nolistsep,noitemsep]
    \item We propose a taxonomy to codify ambiguity in multimodal text-to-code problems;
    \item We propose a number of potential metrics to measure these defined notions of ambiguity, and empirically test which best represents different aspects of ambiguity;
    \item We propose strategies, inspired by theories of discourse, for incorporating dialogue in a coding agent to reduce ambiguity and increase task success.
\end{enumerate}

From our analyses, we find that our metrics can predict ambiguity categories, and pragmatic dialogue increases task success (measured by correctness of the generated code) while targeting the identified ambiguities. We make our code, and annotations publicly available for the camera-ready version of this paper.
\section{Related Work}
\paragraph{Code Generation} Large language models of code have shown strong performance on natural language to code generation \citep[\textit{inter alia}]{Narechania_2021, codex, roziere2024code, lozhkov2024starcoder}. However, work investigating how users interact with code generation models has found that impressive benchmark performance does not always translate to improved task outcomes for users \cite{sarkar2022like, Vaithilingam2022Apr, ma2023ai, mozannar2024realhumaneval, nguyen2024how}. Some of this gap can be attributed to the ambiguity inherent to human interactions with code models: \citet{sarkar2022like} observe that user utterances are often underspecified and ambiguous, forcing users to repeatedly refine their prompts and adapt their thought processes to match the LLM. Likewise, \citet{mozannar2024realhumaneval} observe that users often provide fuzzy instructions, motivating a clarification feature. Underspecified instructions are also present in real-world software engineering benchmarks such as SWE-bench \cite{jimenez2024swebench}, as found during the construction of SWE-bench Verified.\footnote{\url{https://openai.com/index/introducing-swe-bench-verified/}}

Recent work has studied ambiguity resolution for code LLMs via clarification questions. \citet{Mu2024Jul} introduce ClarifyGPT, a pipeline for code generation with selective clarification. 
\citet{li-etal-2023-python} studies clarification for open-domain code generation in a scaffolded setting, and \citet{zhou2025sweetrltrainingmultiturnllm} studies the effects of multi-turn dialogue on programming problems. In addition to \citet{yang2024matplotagentmethodevaluationllmbased}, studying the performance of chat agents on data visualization code. Also, uncertainty estimation and explainability literature have been instrumental for quantifying ambiguity in the works by \citet{Liu2024Jun} and \citet{lin2024generating}. Further, explorations of ambiguity in SQL generation \cite{bhaskar-etal-2023-benchmarking, wang-etal-2023-knowledge} and the role of ambiguity in requirements engineering \cite{Kamsties2005, Bano} have been recently studied.

\paragraph{Ambiguity in NLP Tasks} Ambiguity has been studied across a wide array of NLP tasks, including coreference resolution \cite{poesio-artstein-2005-reliability}, (visual) question answering \cite{min-etal-2020-ambigqa,kim-etal-2023-tree,papakostas-papadopoulou-2023-model,park-etal-2024-picturing} and machine translation \cite{iyer-etal-2023-towards,schouten-etal-2023-reasoning, niwa2024ambignlg, madureira-etal-2024-time}. 
Current language models generally struggle when applied directly to tasks with ambiguity \cite{liu-etal-2023-afraid, zhang-etal-2024-clamber, wildenburg-etal-2024-pre}; by default, they do not recognize ambiguity in instructions, nor do they seek clarification or engage in proactive dialogue to resolve ambiguity \cite{deng-etal-2023-prompting}. However, recent sampling-based methods have shown promise in detecting ambiguity \cite{kadavath2022languagemodelsmostlyknow, kuhn2023semantic, cole-etal-2023-selectively,  lin2024generating}, while prompting and self-improvement methods have proven effective for clarifying ambiguity with LLMs \cite{krasheninnikov2022assistance, kuhn2023clam, andukuri2024stargate}.

\paragraph{Pragmatics}
One approach to resolving ambiguity is to assume the speaker is a rational agent playing a cooperative game \cite{Grice1975Dec} 
where they are choosing an utterance that gives the code generation model the best chance of recovering the program they have in mind. This form of inference has been formalized in the Rational Speech Acts (RSA) framework \citep{frank-goodman-2012-predicting}.
RSA has been productively applied to programming tasks where a user specifies their intent using examples \citep{pu-etal-2020-program,pu2023amortizing,vaduguru2024generating}. Similar approaches to disambiguation also been applied to code generation from natural language using large language models \citep{zhang-etal-2023-coder}. Other pragmatic theories of discourse work include RSA for referential communication in a game of color \cite{monroe-etal-2017-colors, mcmahan-stone-2020-analyzing}, question under discussion \cite{ko-etal-2023-discourse}, and discourse theories as applied to dialogue settings \cite{asher-etal-2016-discourse, chi-rudnicky-2022-structured, atwell-etal-2021-discourse, atwell-etal-2024-combining-discourse, atwell-etal-2022-role}. The frameworks we use to implement our dialogue agents are inspired by these in this work.
\section{Defining and Identifying Ambiguity}
\label{sec:methods}
We define ambiguity to arrive at a taxonomy that helps us identify it automatically. 

\paragraph{Director-Coder Setting}
Initially, we formalize the setting described in \S~\ref{sec:intro}. The \textbf{director} has a target intent $I$ -- a random variable representing the goal image (or corresponding code)\footnote{We do not differentiate between these, since each code corresponds to an image.} -- which the director communicates through a natural language utterance $U(I)$. The utterance $U$ is also a random variable.
The \textbf{coder} interprets this utterance to directly generate code (and corresponding image) $\hat{I} \sim \mathbb{C}(U)$, where $\mathbb{C}(U)$ is a code distribution conditional to the utterance $U$.
We assume there is some evaluation method that produces a random error $E(I, \mathbb{C}(U))$ to quantify goal fulfillment in this context. For example, in later experiments, we use $k$ repeated samples $\hat{I} \sim \mathbb{C}(U)$ and check the accuracy of the code by counting how many pass unit tests (pass@$k$).
In this definition, we will assume that the minimization of the error, $E$, is preferable. 

\subsection{Ambiguity in Plotting Code}
\label{sec:ambiguity}

\paragraph{Definition} We define ambiguity from the perspective of the coder. Intuitively, we frame ambiguity as any portion of uncertainty held by the coder that could be reduced by changing the natural language utterance $U$. More specifically, interpreting $E$ as a form of error, we define ambiguity as the quantity:%
\begin{equation}
\label{eq:def}
\begin{split}
& E(I, \mathbb{C}(U)) - E(I, \mathbb{C}(U^*)) \\
& \text{where} \quad U^* = \text{arg}\min\nolimits_{U' \in \Upsilon} \ E(I, \mathbb{C}(U')),
\end{split}
\end{equation}
and $\Upsilon$ is a constraining set to ensure director utterances are ``reasonable.'' For instance, $\Upsilon$ could be the set of all finite, grammatically correct utterances in the English language.

\paragraph{Interpretation} In a typical machine learning problem, a model $h$ is picked from some constraining set model class $\mathcal{H}$, selecting this model to reduce error as much as possible. For instance, we can select a linear model $h$ from the set $\mathcal{H}$ of all linear models parameterized by elements of $\mathbb{R}^d$. Meanwhile, there is also a best model $h^*$ that minimizes the error for our problem.\footnote{The best model $h^*$ is called the \textit{Bayes optimal} model and the error of this model is the \textit{Bayes error} or the \textit{aleatoric uncertainty} \cite{hullermeier2021aleatoric}.} 
This minimum error (or uncertainty) is simply irreducible without changing the problem definition. Here, we adapt this to define ambiguity. The term $E(I, \mathbb{C}(U^*))$ represents the minimum uncertainty, treating the utterance itself $U$ as the model $h$ we wish to pick, within the aforementioned problem. In turn, ambiguity is formally defined \textit{as the excess error (or uncertainty) of the coder that could have been reduced through improved communication by the director}. This quantity formalizes the process of disambiguation -- studied throughout this paper -- and what it means to disambiguate for a coder with fixed knowledge. 

\begin{figure}[t]
    \centering
    \includegraphics[width=.5\columnwidth]{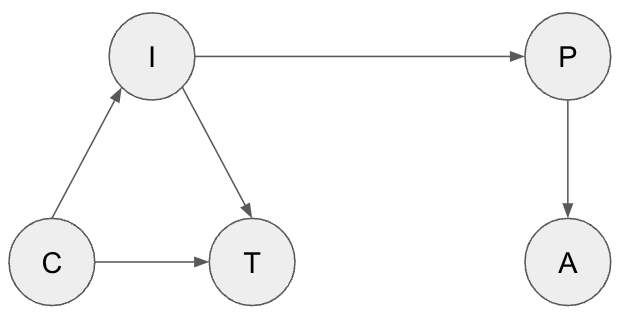}
    \caption{This figure depicts the causal graph of a multimodal coding task. I: Image, C: ground truth code used to generate I, T: unit tests, P: director's prompt, and A: coder's answer. When ambiguity is prevalent, then the edges of this graph are cut, and the coder's answer relies on a non-descriptive prompt.}
    \label{fig:bayesian_graph}
    \vspace{-8pt}
\end{figure}

\paragraph{Taxonomy} Next, we use this definition to arrive at a taxonomy of ambiguities in the plotting domain. We observe that different input modalities to LLMs inherently convey different types of ambiguity (see Figure~\ref{fig:bayesian_graph}), 
as these modalities serve different user intents. For instance, an image in the ``mind's eye'' of the director corresponds to an ideal end product, but the director may not know how this translates to precise code or the best ways to communicate their ideal. This type of cross-modality dependency has been studied by the discourse coherence literature \cite{alikhani-etal-2020-cross, inan-alikhani-2024-seeing}. In order to represent these cross-modal dependencies, we need a taxonomy that captures the different types and sources of ambiguity, which we present next. These categories draw inspiration from cognitive science, linguistics, and discourse literatures, as well as our definition.
\begin{itemize}[nolistsep]
    \item \textbf{Semantic ambiguity:} Certain words and their meanings can have multiple interpretations. Thus, misinterpretation of $U$ by the coder may lead to errors that would be prevented with a better choice of words (i.e., the optimal utterance $U^*$), implying higher ambiguity (Eq.~\ref{eq:def}). This category is based on semantic ambiguity in cognitive science literature as explored by \citet{Zempleni2007Feb, Degani2010Jul}. \\ \textit{e.g. ``regular matplotlib style plot'', ``grouped histogram'', ``color plot''}
    \item \textbf{Presupposition:} Instructions may presuppose default parameter values without explicitly mentioning their use, and disagreement in coder and director presuppositions can also be a source of error caused by the coder's interpretation of $U$, leading to increased ambiguity. This category is based on the presuppositional ambiguity as explored by \citet{2667aa7a-9a7f-3d22-8d50-b0bbb10d809f,Atlas1977Jan,Kempson1979Dec,Jaszczolt2002Jul}. \\
    \textit{e.g. knowing the default parameters of the \texttt{scatter()} function, or original instructions ask to “keep the distribution plot in blue”, but no explicit parameter for “blue” is made because the default is known to be blue.}
    \item \textbf{Underspecification:} Some non-default parameter values or functions are either not mentioned at all, or partially described in the instructions. By leaving out these details, a coder's own interpretation of $U$ can be counter to the director's goals, once again increasing the error compared to $U^*$ and the ambiguity as defined previously. This category is based on the relationship between underspecification and ambiguity as explored by \citet{van1996semantic}. Some context-specific subcategories for scientific plotting can include,
    \begin{itemize}[nolistsep]
        \small
        \item \textbf{Color parameter}, \textit{e.g. “solid red”, alpha value is set but not mentioned}
        \item \textbf{Distance parameter}, \textit{e.g. ``enough'' space between axes}
        \item \textbf{Shape parameter}, \textit{e.g. marker shape is set even though not mentioned}
        \item \textbf{Size parameter}, \textit{ e.g. marker size is set even though not mentioned}
        \item \textbf{Location parameter}, \textit{e.g. legend location is set even though not mentioned}
        \item \textbf{Label parameter}, \textit{e.g. title is set to be “xxx” even though not mentioned}
        \item \textbf{Line parameter}, \textit{e.g. ``full line'', ``dashed line''}
        \item \textbf{Function}, \textit{e.g. to plot a heatmap, using either \texttt{imshow()} or \texttt{pcolor()}.}
    \end{itemize}
\end{itemize}

\begin{figure*}[!t]
    \centering
    \includegraphics[width=.8\textwidth]{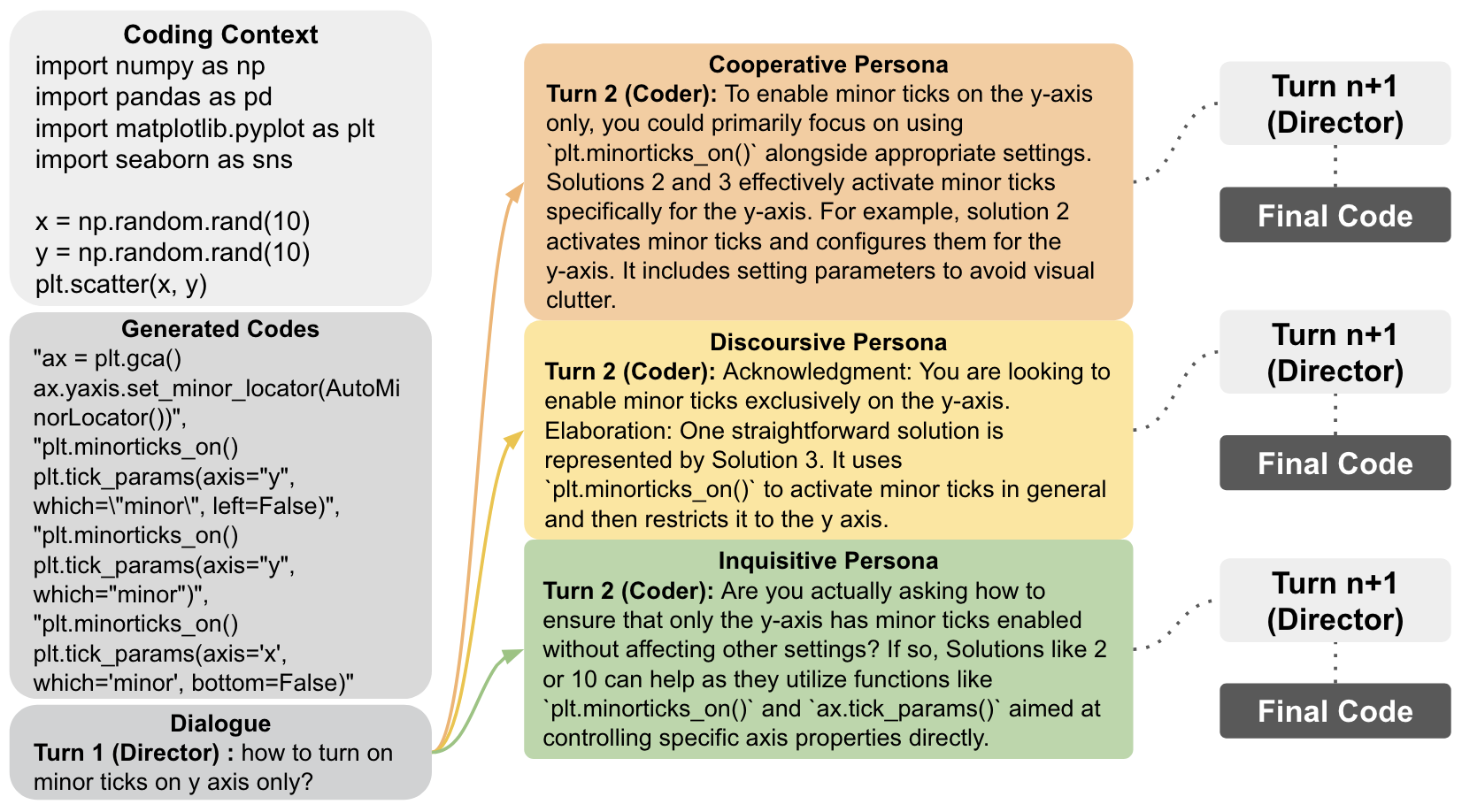}
    \caption{This figure shows the dialogue flow for a pragmatic coder, where the initial intent of the dialogue is given on the left, and the different responses generated using separate personas are given in the middle.}
    \label{fig:dialogue_flow}
    \vspace{-5pt}
\end{figure*}

\paragraph{Annotating for Ambiguity}
We carry out an annotation and do a preliminary analysis of the distribution of these categories of ambiguity in the DS1000 dataset \cite{lai2022ds1000}, specifically with questions from plotting libraries (e.g., \texttt{Matplotlib}, \texttt{Seaborn}) \cite{Hunter:2007}. This dataset, featuring natural language prompts from StackOverflow with human-written tests and plots, inherently contains potential human-introduced ambiguities.
We, the authors, who have prior experience with ambiguity in dialogue, annotated 155 coding instructions from the DS1000 dataset. Despite the DS1000 dataset's claim that the problems are written to be unambiguous, we still find that 57\% of the plotting questions fall under one or more of the categories we have defined above. Interannotator agreement on 25 sampled questions, measured by Cohen's Kappa ($\kappa=0.587$), indicates moderate to substantial agreement, as anticipated for this subjective meta-labeling task \cite{Fleiss1973Oct, Banerjee1999Mar, viera2005understanding}. Table~\ref{tab:ds1000_amb_stats} shows the distribution of different categories.

\begin{table}[!h]
    \centering
    \resizebox{0.75\columnwidth}{!}{
    \begin{tabular}{c|c}
        \toprule
        \textbf{Ambiguity Category} & \textbf{Distribution} \\
        \midrule
        semantic ambiguity & 23.8\% \\
        presupposition & 11.9\% \\
        underspecification & 73.8\% \\
        \bottomrule
    \end{tabular}
    }
    \caption{This table shows the ambiguity category distribution within the ambiguous prompts (57\%) of the 155 plotting questions in the DS1000 dataset based on our annotations.}
    \label{tab:ds1000_amb_stats}
    \vspace{-10pt}
\end{table}


\subsection{Automatic Metrics for Ambiguity}
In addition to the human annotations, we propose automatic ways of measuring ambiguity based on our formalizations and taxonomy (\S\ref{sec:ambiguity}), and compare them to traditionally employed uncertainty-based metrics.

\paragraph{Sampling Diversity (SD)} A common approach \cite{cole-etal-2023-selectively} to measuring uncertainty about a given utterance \(U \) is to count the number of distinct programs that satisfy the constraints specified in the prompt. If the coder's solution distribution $\mathbb{C}(U)$ assigns non-zero probability to many codes (i.e., it generates many different codes upon sampling), then these codes all differ in apsects unconstrained by the prompt $U$ from the coder's perspective.
This indicates the coder is uncertain about the intent of $U$. The more distinct programs there are for a given sample size, the higher the coder's uncertainty about the utterance is. This method serves as a baseline for ambiguity measurement; while it does capture uncertainty about the utterance $U$, it does not capture information about the optimal prompt $U^*$ in any way. This is an important distinction between typical notions of uncertainty and our proposed definition of ambiguity. We give the implementation details in Appendix~\S\ref{sec:appendix_AST}, where we compare Abstract Syntax Trees (AST) of both programs using edit distance.

\paragraph{Repeated Parameter Counting (RPC)} Another way to quantify ambiguity in an utterance \( U \) is to focus on identifying function calls/parameters that appear across the set of proposed programs. This is conceptually similar to sampling diversity, which compares distinct solutions directly. We hypothesize function calls and parameters may better capture presuppositional ambiguity within our taxonomy since this ambiguity is directly related to the default parameter settings within a code library. More specifically, our proposed RPC metric measures ambiguity by counting which function calls and parameters remain constant across the code solutions. The fewer elements that must remain fixed, the higher the ambiguity. Similar to sample diversity, this metric does not explicitly consider the optimal prompt $U^*$. We expect it to perform well for presupposition, because it focuses on aspects of code that are often presupposed.

\paragraph{Optimal Result Gap (ORG)} Building on our definition in Eq.~\ref{eq:def}, we propose a subfamily of metrics that attempt to directly compute this mathematical notion of ambiguity. While the coder's realized uncertainty $E(I, \mathbb{C}(U))$ is easy to compute by using Pass@$k$ scores for a given utterance $U$, the 
minimum uncertainty $E(I, \mathbb{C}(U^*))$ is more difficult because we cannot be sure of the optimal utterance $U^*$. We propose to approximate $U^*$ by using one of three oracle information sources: ground-truth code for the plot, ground-truth image for the plot, and unit tests for the ``Pass'' determination. We provide these oracle sources to a large language model and ask it to provide a prompt that enable itself to generate the code, create the image with code, or generate code that would pass the tests. While this approach provides an estimate, we acknowledge its limitations: the model-generated re-prompt may not fully capture an ideal, ambiguity-free prompt, making it an imperfect but practical approximation of the optimal formulation. At the same time, we expect it to improve over the previous two metrics because it does consider the optimal prompt $U^*$.

\paragraph{Self Verification} This is a traditional method of uncertainty estimation where a model is asked to return its uncertainty about a solution when provided a problem statement. This baseline uncertainty methodology, like sampling diversity and RPC, does not explicitly consider the optimal utterance $U^*$, which is an important aspect of our definition. We use a prompt similar to the ones used by \citet{cole-etal-2023-selectively} and \citet{sicilia2024accountingsycophancylanguagemodel}. 

\paragraph{LLM-Based Ambiguity Rating (LAR)} On the other hand, we can extend self-verification to better consider our definitions of ambiguity. Instead of querying the model for its uncertainty, we can prompt it to rate the ambiguity of \( U \) on a scale of 1 to 10, providing the model with our ambiguity taxonomy as a resource in the prompt. This encourages the rating aligning with our pre-defined ambiguity categories, rather than being an arbitrary self-assessment. Since these categories are also based on our initial mathematical definition of ambiguity, it also  serves as heuristic approximation of Eq.~\ref{eq:def}. Different from baseline methods focused on uncertainty, it implicitly considers the optimal prompt $U^*$ through the characterizations of optimality encoded in our taxonomy.
\section{Disambiguation with Coding Dialogues}
\label{sec:dialogue}
After formalizing ways of identifying ambiguity, we now propose that ambiguity can be resolved using dialogues, and we formalize a dialogue setup with persona-based generation components.

\subsection{Basic Dialogue Setup}
We propose resolving ambiguity in natural language specifications of intent with multi-turn dialogue. Each coding task is defined by the natural language intent $I$ (see Figure~\ref{fig:dialogue_flow}) and the initial request $U_1$, as before. 
%
Meanwhile, our proposal extends the previous setting to allow for dialogue:
\begin{enumerate}[noitemsep,nolistsep]
    \item Director presents instruction $U_1(I)$.
    \item Coder responds with an utterance $U_2(U_1)$.
    \item Director continues $U_i(U_{:i-1}, I)$, using access to target image and utterance history.
    \item Coder continues $U_{i+1}(U_{:i-1})$, using access to utterance history only.
    \item Repeat steps 3 and 4 based on the number of turns controlled as a hyperparameter.
    \item Session always terminates with coder providing their final code $\hat{I}(U_{:})$.
\end{enumerate}
In this setting, coder utterances can include dialogue acts like clarification questions, e.g., ``C: what location should I put the legend,'' which evokes a specific response, ``D: The top left corner'' or can be more general declarations that start a sub-topic of conversation ``C: I'll plan for the default legend arguments. D: Hmm. Keep it on the top left. What else can you change?''


\subsection{A Pragmatic Dialogue Setup}
\label{sec:frameworks}
We hypothesize ambiguity reduction in our proposed dialogue setting can be improved by encouraging LLM-based coders to consider pragmatics in their dialogue strategy. We operationalize this by using persona prompting and in-context learning as described in \citet{Wang2023Oct, schulhoff2024prompt}, and \citet{zheng2023a}. 
Next, we describe the personas we use to generate responses. These are based on three theories of discourse: cooperative, discoursive, and inquisitive. For the implementation details of this setup, please refer to Appendix~\ref{sec:appendix_dialogue_gen}.
\subsubsection{Dialogue Strategies}
\paragraph{Cooperative}
The first framework we use is based on Grice's maxims of cooperative dialogue partners \cite{Grice1975Dec, horn1984, Levinson2000Apr, mcmahan-stone-2020-analyzing}. Here, the coder is a pragmatic agent that recursively engages in interaction and models the director's state of mind to respond to an utterance. We use the Gricean cooperativity principle to design the prompt for this dialogue strategy, given in Appendix~\ref{sec:appendix_prompting_details}.
\paragraph{Discoursive}
The second pragmatics framework is based on Discourse theories. Here, the coder is not necessarily responding strategically. Still, it's utterance is always related to the history of the conversation and the coding context through a set of coherence relations. Hence, when a coder produces an utterance, it relates to the set of solutions it has sampled as well as what the director has said in the previous turn. This definition of discourse is mostly similar to SDRT-like
dialogue-based relation categories \cite{ko-etal-2023-discourse, asher-etal-2016-discourse, fu-2022-towards, atwell-etal-2024-combining-discourse, Alikhani2023May}. We provide the persona prompt in Appendix~\ref{sec:appendix_prompting_details}.
\paragraph{Inquisitive}
The third pragmatics framework is related to discourse theories, but focuses more on question-type relations. In this case, the coders utterance explicitly answers an implicit question posed by the director. This discourse framing is described by \citet{Clifton2012Sep} under the umbrella term of Questions Under Discussion (QUD). When a director gives a coding instruction, the pragmatic coder with QUD understanding first detects an implied question indirectly posed by the director and the coder answers that question. 
The persona prompt is given in Appendix~\ref{sec:appendix_prompting_details}. 
\section{Experiments \& Findings}
\label{sec:experiments}
In this section, we provide details of our experiments of disambiguation, and user simulation and show the utility of our ambiguity taxonomy and metrics. We answer multiple research questions and report our findings in combination with our experiments. We first describe the experiments for our taxonomy and automatic metrics as described in Section~\S\ref{sec:ambiguity}, and then follow up with experimentation based on the dialogue approach to coding we described in Section~\S\ref{sec:dialogue}. We experiment mainly with GPT-4o in our experiments, but provide additional results for LLaMA-3.2, StarCoder, and CodeLLaMA in Appendix~\S\ref{sec:appendix_different_models}.
\begin{table}
\centering
\resizebox{.9\columnwidth}{!}{%
\begin{tabular}{@{}lccc|c}
\toprule
 &
  \multicolumn{1}{c}{\textbf{Sem. Amb.}} &
  \multicolumn{1}{c}{\textbf{Underspec.}} &
  \multicolumn{1}{c}{\textbf{Presup.}} &
  \multicolumn{1}{|c}{\textbf{Avg.}} \\ \midrule
RPC &
  \cellcolor[HTML]{FFFFFF}0.450 &
  \cellcolor[HTML]{FFFFFF}0.412 &
  \cellcolor[HTML]{FFFFFF}0.466 &
  \cellcolor[HTML]{FFFFFF}0.443 \\
ORG$_C$ &
  \cellcolor[HTML]{E3F4EB}0.527 &
  \cellcolor[HTML]{FFFFFF}0.495 &
  \cellcolor[HTML]{FFFFFF}0.326 &
  \cellcolor[HTML]{FFFFFF}0.449 \\
ORG$_I$ &
  \cellcolor[HTML]{97D5B7}0.597 &
  \cellcolor[HTML]{FFFFFF}0.450 &
  \cellcolor[HTML]{FFFFFF}0.451 &
  \cellcolor[HTML]{FFFFFF}0.499 \\
ORG$_U$ &
  \cellcolor[HTML]{BDE5D1}0.561 &
  \cellcolor[HTML]{FFFFFF}0.494 &
  \cellcolor[HTML]{FFFFFF}0.445 &
  \cellcolor[HTML]{FFFFFF}0.500 \\
LAR &
  \cellcolor[HTML]{57BB8A}0.655 &
  \cellcolor[HTML]{FFFFFF}0.453 &
  \cellcolor[HTML]{FFFFFF}0.447 &
  \cellcolor[HTML]{ECF7F2}0.518 \\
LAR$_\mathcal{T}$ &
  \cellcolor[HTML]{A4DBC0}0.585 & 
  \cellcolor[HTML]{DFF2E9}0.530 & 
  \cellcolor[HTML]{7CCAA4}0.622 & 
  \cellcolor[HTML]{AADDC4}0.579 \\
SV &
  \cellcolor[HTML]{FFFFFF}0.380 &
  \cellcolor[HTML]{FFFFFF}0.399 &
  \cellcolor[HTML]{FFFFFF}0.493 &
  \cellcolor[HTML]{FFFFFF}0.424 \\
  \bottomrule
\end{tabular}%
}
\caption{This table shows the AUC scores between different ambiguity metrics we propose and the ambiguity categories from our taxonomy. This shows that most metrics are predictive of semantic ambiguity, while underspecification and presupposition are less correlated. Subscripts indicate the ground truth modality: I (images), C (code), U (unit test).}
\label{tab:auc_metrics}
\vspace{-10pt}
\end{table}

\paragraph{Our Metrics Predict Ambiguity Categories} To test the hypothesis of whether our metrics are predictive of ambiguity (as we defined), we carry out a correlation study using ROC AUC scores\footnote{Area Under the ROC Curve (\textbf{AUC}) is a measure of correlation about how well a continuous independent variable can predict a binary dependent variable, via testing a variety of different thresholds. An AUC of 0.5 is equivalent to a random baseline, while an optimal score is 1.} (Table~\ref{tab:auc_metrics}). In our case, we use it to measure correlation between the ambiguity scores and the ground-truth human-produced ambiguity labels for each instance of the 155 plotting questions. We observe that ORG, which contains the oracle solutions to approximate ambiguity, predicts the semantic ambiguity category, but less so the other categories. This may be due to the class imbalance in the dataset, and also the suboptimal nature of reprompts used in the ORG metrics. The most predictive of any ambiguity category is the LAR$_\mathcal{T}$ metric, where a model is asked to rate the ambiguity using our taxonomy. This shows the validity and applicability of this metric to unsupervised contexts. The lowest prediction power comes from the traditional uncertainty measurement technique of self-verification (SV), as it does not necessarily correlate with the ambiguity of the user's intent, but the uncertainty of the model providing an answer to the prompt.  

\begin{table}[!t]
\centering
\resizebox{.8\columnwidth}{!}{%
\begin{tabular}{cl|c}
\toprule
\multicolumn{2}{l}{{\color[HTML]{000000} }}                       & \textbf{Pass@1} \\ \midrule
\multicolumn{2}{c|}{{\color[HTML]{000000} Baseline (no dialogue)}} & 68.38\%     \\ \midrule
{\color[HTML]{000000} }   & \cellcolor[HTML]{FFFFFF}Cooperative   & 79.44\%     \\
{\color[HTML]{000000} }   & \cellcolor[HTML]{FFFFFF}Discoursive   & 74.11\%     \\
\multirow{-3}{*}{{\color[HTML]{000000} \begin{tabular}[c]{@{}c@{}}With Reference \\ Code\end{tabular}}}  & \cellcolor[HTML]{FFFFFF}Inquisitive & 66.34\% \\ \midrule
{\color[HTML]{000000} }   & \cellcolor[HTML]{FFFFFF}Cooperative   & 75.23\%     \\
{\color[HTML]{000000} }   & \cellcolor[HTML]{FFFFFF}Discoursive   & 74.06\%     \\
\multirow{-3}{*}{{\color[HTML]{000000} \begin{tabular}[c]{@{}c@{}}With Reference \\ Image\end{tabular}}} & \cellcolor[HTML]{FFFFFF}Inquisitive & 64.56\% \\ \midrule
\multicolumn{2}{c|}{{\color[HTML]{000000} \begin{tabular}[c]{c}Ceiling Performance \\ (Non-Ambiguous Reprompt)\end{tabular}}}  & 87.74\% \\ \bottomrule
\end{tabular}%
}
\caption{This table presents the mean pass@1 scores for different types of dialogue strategies that we propose (\S\ref{sec:dialogue}). The baseline corresponds to the GPT-4o code answers to the original prompts, while the ceiling performance uses non-ambiguous reprompts. Having a dialogue with pragmatics-inspired personas improves task success drastically, yet there is still ambiguity between the ceiling performance.}
\label{tab:pragmatics_results}
\vspace{-10pt}
\end{table}
\begin{table*}[!ht]
\centering
\resizebox{\linewidth}{!}{
\begin{tabular}{@{}cccccc@{}}
\toprule
\textbf{Coding Question} &
  \textbf{Ambiguity} &
  \textbf{Baseline} &
  \textbf{Cooperative} & \textbf{Discoursive} & \textbf{Inquisitive} \\ \midrule
\begin{tabular}[c]{@{}c@{}}draw a line (with random y) for each \\ different line style\end{tabular} &
  underspecification &
  0.000 &
  0.000 &
  0.200 &
  0.000 \\
draw a full line from (0,0) to (1,2) &
  semantic ambiguity &
  0.000 &
  0.067 &
  0.000 &
  0.000 \\
\begin{tabular}[c]{@{}c@{}}make seaborn relation plot and color \\ by the gender field of the dataframe df\end{tabular} &
  underspecification &
  0.067 &
  0.533 &
  0.000 &
  0.000 \\
highlight in red the x range 2 to 4 &
  semantic ambiguity &
  0.667 &
  0.967 &
  1.000 &
  0.167 \\ \bottomrule
\end{tabular}
}
\caption{This table shows a breakdown of the final executability scores (pass@1 with 30 samples each instance) for different questions in the DS1000 dataset, with their annotated ambiguity categories. The examples are picked to show when most models have low scores, or to show the performance according to different categories of ambiguity. }
\vspace{-7pt}
\label{tab:error_analysis}
\end{table*}

\paragraph{Pragmatic Dialogue Increases Task Success} To test the hypothesis of whether the pragmatic dialogue setup that we proposed in \S\ref{sec:dialogue} disambiguates and improves task success (as measured by the pass@$k$ correctness score), we carry out a comparative experiment with results shown in Table~\ref{tab:pragmatics_results}. Here, we test two scenarios, one in which the director is given the reference code and one where the reference image is used. This comparison alleviates the concern about whether there is ground truth code leakage from the director to the coder. It can be observed that the best-performing dialogue strategy is pragmatic cooperative reasoning in both categories, likely benefiting from the theory-of-mind reasoning and chain-of-thought training in modern LLMs. 

The inquisitive strategy is the least-performing model, even worse than the baseline, which may mean that always looking for questions under the discussion can hinder disambiguation. Despite improvements, a gap remains between the best-performing strategy and the ceiling performance, indicating unresolved ambiguity. Since user intent is fixed in this static task, full resolution is unlikely, and even the ceiling performance is imperfect, as the re-prompt itself may still contain ambiguity.

\begin{figure}[!ht]
    \centering
    \includegraphics[width=\columnwidth]{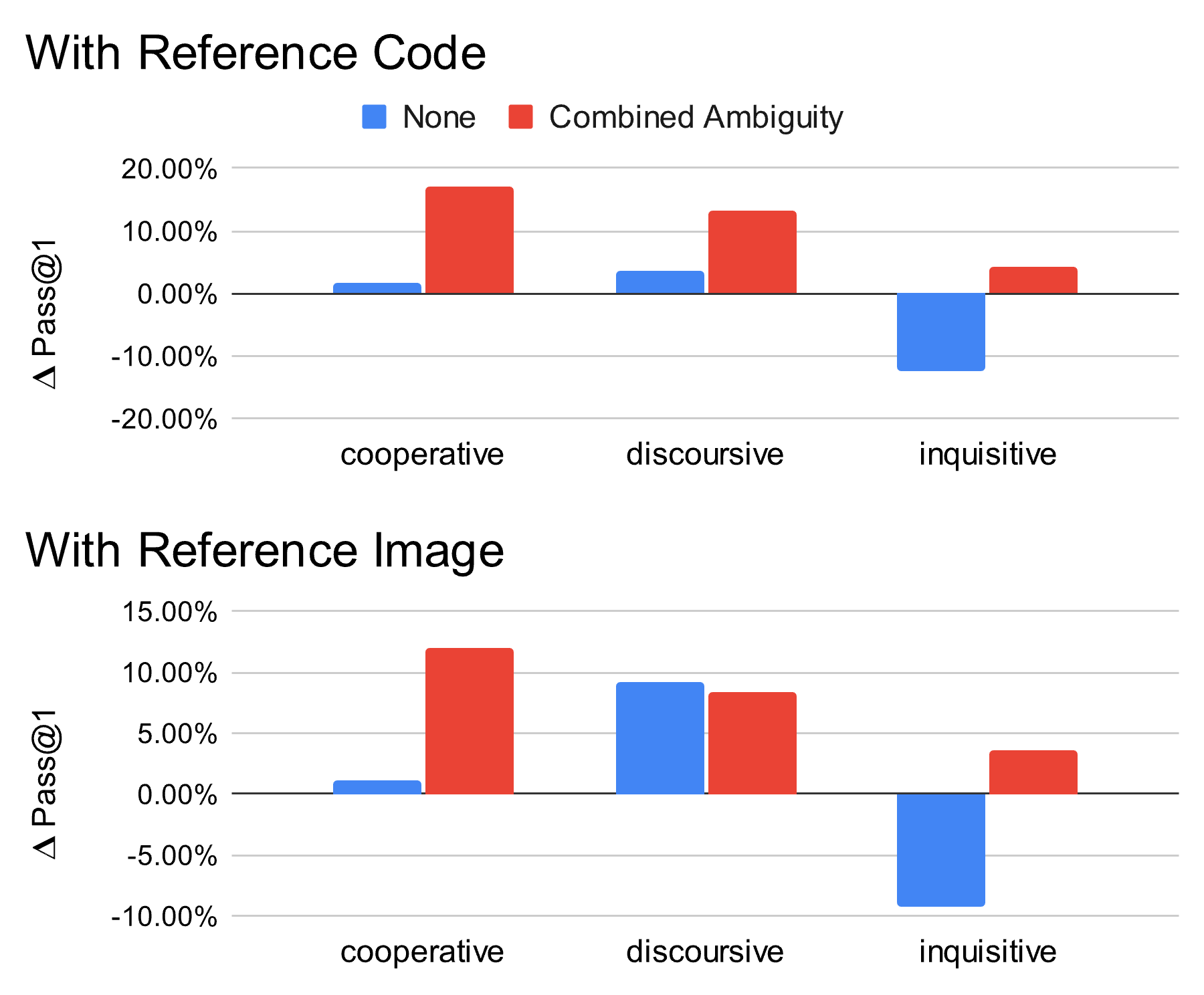}
    \caption{This figure shows a breakdown of the change in the mean pass@1 scores ($\Delta$ = post-dialogue $-$ original) across ambiguous and non-ambiguous instances of the DS1000 dataset. Dialogue shows better performance in ambiguous instances instead of non-ambiguous ones.}
    \label{fig:delta_pass_at_1}
    \vspace{-7pt}
\end{figure}

\paragraph{Pragmatic Dialogue Targets Ambiguities}  
To evaluate whether dialogue strategies improve code accuracy by directly addressing the ambiguities identified in our taxonomy, we measure the change in mean pass@1 scores between ambiguous and non-ambiguous cases (Figure~\ref{fig:delta_pass_at_1})\footnote{We focus specifically on task success rather than user satisfaction due to the subjectivity and costs of user experiments, while already showing that ambiguity is addressed.}. The results clearly show that dialogue-driven improvements are consistently greater for ambiguous cases than for non-ambiguous ones. This confirms that dialogue effectively disambiguates prompts across all three pragmatic personas. However, when the reference image is provided instead of the code, the Discoursive Persona performs similarly in both cases, suggesting that dialogue alone may not fully resolve ambiguities. Additionally, in line with Table~\ref{tab:pragmatics_results}, $\Delta$pass@1 for the Inquisitive Persona is negative in non-ambiguous instances, yet it still succeeds in clarifying ambiguous prompts.

%
\subsection{Error Analysis}
\label{sec:error_analysis}
Table~\ref{tab:error_analysis} presents a detailed performance breakdown of different dialogue strategies. Notably, certain questions remain challenging even after dialogue, yet specific ambiguity categories align with the most effective pragmatic strategy. For instance, nearly all personas failed to resolve the first underspecification question (mean pass@1: 0.000), with only the cooperative persona achieving occasional success (mean pass@1: 0.267). Interestingly, in some cases, additional dialogue negatively impacted performance across all personas. The inquisitive persona performed best for vagueness-related ambiguities, while the discoursive and cooperative personas excelled in addressing parameter underspecification.
\section{Conclusion}
Pragmatic theories of language emphasize that meaning is never fully encoded in words alone but emerges through the interactional work of negotiating intent. Drawing on this perspective, our study shows how ambiguity in natural language code specifications, whether through underspecification, presupposition, or semantic vagueness, can be systematically identified and quantified. By translating concepts from discourse and linguistic theory into a taxonomy and set of operational metrics, we move beyond ad hoc uncertainty measures toward a principled account of when and why user intent becomes opaque. This reframing positions ambiguity not as incidental error but as an analyzable property of communication between humans and coding agents.

Building on this foundation, we demonstrate that dialogue offers a practical path for resolving these ambiguities. We characterized various pragmatics frameworks in relation to pair-programming-like dialogues that happen between a director and a coder. We then analyzed the effects of having dialogues with different reasoning strategies on the executability and disambiguation of the final generated code. As having a dialogue based on code is becoming the norm with LLMs, focusing on the pragmatics of dialogue opens up new venues for developing dialogue systems, datasets, and evaluation mechanisms for code generation. With this, future coding assistants can transform ambiguity into a space for alignment, producing collaborations that are both more accurate and more human-like.

\section*{Limitations}
We proposed using pragmatic dialogue for code generation, but the major limitation is from the side of human data collection and evaluation. We resorted to automatic metrics already being used or developed for this study to evaluate our setup without relying on human annotators. However, this entails that the evaluations may not be human-like and may not show the most accurate representations even though they show improvements in generally accepted code executability standards. Further, we did not deploy a dialogue system to study our approach. Instead, we resorted to simulations using LLMs, which may or may not accurately represent how a human interlocutor would act in a real-world setting. We wanted to minimize this by using large parameter models for dialogue generation and StackOverflow-based code instructions from the DS1000 dataset.

\section*{Ethics Statement}
In our simulation process we have used GPT-4o, and this is a closed-source LLM, and we are aware that this model can propagate its own training biases. The scientific community does not have access to any information regarding how this model is trained or what the dataset consists of. This may result in a deficient evaluation of the final performance and human-likeness of the generated dialogue. This is a simulated analysis study to identify and characterize pragmatics frameworks with possible LLM behavior in a pair programming setting. Hence, we do not involve humans in our current setup. The biases propagated by GPT-4o are the responsibility of OpenAI and should be held accountable by their and the scientific community's ethical standards.


\bibliography{custom}

\appendix

\section{Generating Dialogue Responses}
\label{sec:appendix_dialogue_gen}
We simulate our dialogues for plotting code using LLMs and based on the algorithm given in Algorithm~\ref{alg:dialogue_sim}. In the algorithm, $f_D$ and $f_C$ are defined based on different pragmatics strategies as given in detail in Section \S\ref{sec:frameworks}.

\begin{algorithm}[!h]
\begin{algorithmic}[1]
\caption{Dialogue Simulation with LLMs.}\label{alg:dialogue_sim} 
\REQUIRE Problem instance $(u^D_1, I)$
\REQUIRE Director model $f_D$
\REQUIRE Coder model for NL response $f_C$
\REQUIRE Coder model to generate code $g_C$
\REQUIRE Number of samples $k$
\REQUIRE Number of rounds of dialogue $n$
\STATE $S \gets \{s_i \sim g_C(u^D_1)\ |\ 1 \leq i \le k\}$
\STATE $\mathbf{u} \gets [\ ]$
\FOR {$n$ times}
    \STATE $u^C\gets f_C(\mathbf{u})$
    \STATE $\mathbf{u} \gets \mathbf{u} + [u^C]$
    \STATE $u^D \gets f_D(I, \mathbf{u})$  
    \STATE $\mathbf{u} \gets \mathbf{u} + [u^D]$
\ENDFOR
\STATE $c \sim g_C(\mathbf{u})$
\RETURN $\mathbf{u}$, $c$
\end{algorithmic}
\end{algorithm}

\paragraph{Director}
We prompt the director model $f_D$ to generate instructions and clarifications that guide a coder model toward the correct solution. Since we work with an artificial director agent, we source intents from the DS-1000 dataset. We present the intent to the director in one of two ways -- as the code for a reference solution or the plot generated by the code presented as an image. Since a natural language instruction accompanies the DS-1000 problem instances, we seed the interaction using that interaction as the first director turn ($u^D_1$). We prompt the model to use different strategies to generate responses.

        
\paragraph{Pragmatic Coder}
We first extract the code context and the coding instructions from the DS1000 dataset and then convert it into a dialogue format as described in section \S\ref{sec:frameworks}. Then, using GPT-4o, we generate codes that respond to the original instruction (sampled k times). To the pragmatic coder, we present a set of possible unique answers it can choose from the generated codes and the dialogue history that is happening and ask for a follow-up utterance for the coder to converge to the solution that the director is describing, i.e. $g_C(u^D_1)$. We then instruct it to give three solutions based on the reasoning types. For the regular director, we provide the reference code (or the reference plot in the case of a multimodal model) and the dialogue history and ask to generate a follow-up utterance to converge to a solution without giving away the answer. All the details of the prompts are given in Appendix~\ref{sec:appendix_prompting_details}.

\paragraph{Dialogue Policy} We employ a rule-based dialogue policy to choose one of the three utterances we generated for each strategy in the simulation. 
For the first turn of the dialogue, we do not use any LLM generations but directly use the coding instruction from the DS1000 dataset. For the following turns, we generate three different utterances, one for each of the pragmatic director's reasoning ways, and then generate a single utterance without any pragmatic reasoning prompting for the coder for each of the three responses of the director.
We use the number of turns as a hyperparameter to generate the dialogue and perform ablation experiments on it. We do not mix reasoning styles across the dialogue's turns, but we choose a single reasoning style for the overall dialogue. We also experiment with providing the reference image or the reference code to the director to see how clarity of instructions affects execution.

\section{Prompting Details}
\label{sec:appendix_prompting_details}

\begin{figure*}
    \centering
    \begin{tabular}{ccc}
        \includegraphics[width=.3\linewidth]{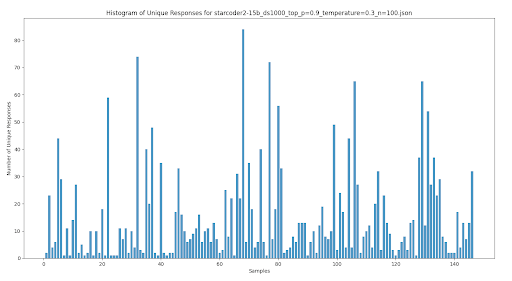} & \includegraphics[width=.3\linewidth]{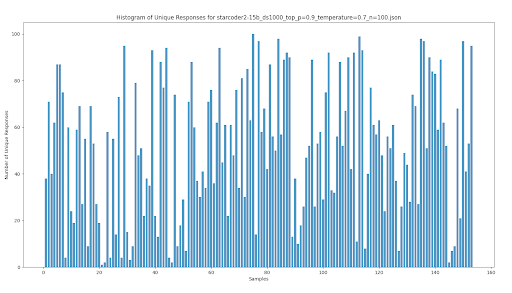}  & \includegraphics[width=.3\linewidth]{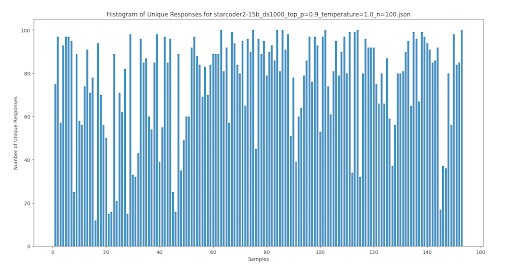}
    \end{tabular}
    
    \caption{This figure shows the change in unique responses of code completions depending on the temperature of the model. From the left, the plots are showing histograms for 0.3, 0.7, and 1.0 temperatures. The horizontal axis is the question number from the DS1000 matplotlib dataset. It is observable that the uniqueness is high for higher temperatures, expectedly. However, very high temperatures may have minor differences that increase the overall uniqueness. Hence, a moderate temperature like 0.7 gives more reliable results for further experimentation.}
    \label{fig:temp}
\end{figure*}

\subsection{Pragmatic Coder}

\subsubsection{System Prompts:}

\textbf{Director:}
You are a coding director. There is another coding agent you are going to have a dialogue with. You have a final product in mind. This is going to be named the REF CODE. You want a coder to write the codes for this final product. For the first turn of the dialogue, you give a specific instruction or a question about the final product. Then, the coder will give you some answers, and then you will have another turn to refine the codes.

\textbf{Coder:}
You are a coding agent. There is another director agent you are going to have a dialogue with. The things you say depend on your persona. You have the following different personas (reasoning styles):

- Cooperative Persona (Pragmatic): You want to converge on the solution as quickly as possible and follow Grice's Maxims when choosing your words. You anticipate the director's cooperative reasoning. You possess theory-of-mind capabilities and common sense. 

- Discourse Reasoning Persona: Everything you say is connected to the previous turn with a relation. The possible discourse relations are Comment, Clarification Question, Elaboration, Acknowledgment, Continuation, Explanation, Conditional, Alternation, Result, Background, Narration, Correction, Parallel, Contrast. You try to identify the relation between the utterance of the director in the previous with your utterance. Then you reply with an utterance that has the appropriate relation. 

- Questioning Persona: Everything you say has an implicit question underneath it. You should tell what the director is actually asking for (the question under their instruction), and give your answer to that implicit question.

The director has a final product in mind. You, as the coder, write the codes for this final product or have a dialogue about the instruction. For the first turn of the dialogue, the director gives a specific instruction or a question about the final product. Then, you will give some answers, and then the director will have another turn to refine the codes.

\textbf{user prompts:}

\textbf{Director:}
REF CODE: ```+ ref-code ``` + DIALOGUE HISTORY:" + dialogue-history + What can you say on the follow-up turn for the coder to converge to the reference code? Do not mention anything about the REF CODE, and don't give away the answer.

\textbf{Coder:}
POSSIBLE GENERATED CODES:
Solution 1: ```CODE```
Solution 2: ```CODE```
....

DIALOGUE HISTORY: + dialogue-history + What can you say on the following turn as the coder to converge to the solution that the director has in mind? Give responses for all types of your personas. Personas must not give the same solution! Your solution MUST NOT contain any new code. You can talk about the provided code.

\section{Additional Experiments with Various Models}
\label{sec:appendix_different_models}
This section presents results from several experiments with multiple other models, such as CodeLLaMA, LLaMMA3.2, StarCoder-2 in Table~\ref{tab:other_model_results} and Table~\ref{tab:gpt_llama_results}. 

\begin{table}[!h]
\resizebox{\columnwidth}{!}{
\begin{tabular}{@{}lllcc@{}}
\toprule
 &
   &
   &
  \multicolumn{1}{l}{\textbf{Pass@1}$\uparrow$} &
  \multicolumn{1}{l}{\textbf{$d_s$}$\downarrow$} \\ \midrule
\multicolumn{2}{c}{Baseline}             & No Dialogue      & 0.422 & 0.744 \\ 
\midrule
\multirow{6}{*}{\begin{tabular}{c} Pragmatic \\ Coder \end{tabular}} &
  \multirow{3}{*}{with code} &
  Cooperative  &
  0.427 &
  0.640 \\
 &                                       & Discoursive    & 0.467 & 0.613 \\
 &                                       & Inquisitive  & 0.396 & 0.716 \\
 & \multirow{3}{*}{with image} & Cooperative  & 0.447 & \textbf{0.584} \\
 &                                       & Discoursive    & \textbf{0.493} & 0.624 \\
 &                                       & Inquisitive  & 0.393 & 0.711 \\ \bottomrule
\end{tabular}
}
\caption{This figure shows the main results of our experimentation for CodeLLaMA as the coder, and GPT-4o as the director, and the baseline corresponds to StarCoder2. Here, we give the metrics for both executability and sampling diversity. Having a dialogue generally performs better than the baseline code completion without any dialogue. For each pragmatic setting, we experiment with all the reasoning styles and have an image or code as the reference solution for the director.}
\label{tab:other_model_results}
\end{table}

\begin{table}
\centering
\resizebox{\columnwidth}{!}{%
\begin{tabular}{@{}llrrr@{}}
\toprule
\textbf{}         & \multicolumn{4}{c}{\textbf{No Dialogue}}                                                                                       \\ \midrule
\textbf{Pass @ 1} & \textbf{OG}                 & \multicolumn{1}{l}{\textbf{I}} & \multicolumn{1}{l}{\textbf{C}} & \multicolumn{1}{l}{\textbf{U}} \\ \midrule
GPT 4o            & \multicolumn{1}{r}{68.38\%} & 68.38\%                        & 87.74\%                        & 81.29\%                        \\
LLaMA 3.2         & \multicolumn{1}{r}{63.23\%} & 64.52\%                        & 77.42\%                        & 65.81\%                        \\
LLaMA → GPT       & -                           & 66.45\%                        & 85.16\%                        & 78.71\%                        \\
GPT → LLaMA       & -                           & 77.42\%                        & 81.29\%                        & 79.36\%                        \\ \bottomrule
\end{tabular}%
}
\caption{This table shows the results for ambiguity representation transfer between different models. OG: original prompt, I: image reprompt, C: code reprompt, U: unit test reprompt. }
\label{tab:gpt_llama_results}
\end{table}

\section{Abstract Syntax Tree (AST) Functional Uniqueness Algorithm}
\label{sec:appendix_AST}
In this section, we detail the AST-based function uniqueness comparison algorithm between two separate generated functions. The code for the algorithm is given in Listing \ref{lst:ast_comp}. We find this form of comparison to be appropriate for plotting tasks as the lines of code of interest are generally the calls to library functions, particularly those provided by the \texttt{matplotlib} API.

\begin{figure*}[t]
\begin{lstlisting}[language=Python, caption=This code snippet shows how the functions of two separate generated codes are compared using their ASTs., label=lst:ast_comp, numbers=none]
def compare_parse_trees(response1, response2):
    """Compare the parse trees of two responses."""
    unique_function_calls = []
    unique_params = {}
    unique_keywords = {}
    try:
        tree1 = ast.parse(response1)
        functions1 = get_params(tree1)
        tree2 = ast.parse(response2)
        functions2 = get_params(tree2)
        for function in functions1.keys():
            if function not in functions2.keys():
                unique_function_calls.append(function)
            else:
                for i, arg in enumerate(functions1[function]):
                    if arg not in functions2[function]:
                        if function not in unique_params.keys():
                            unique_params[function] = []
                        unique_params[function].append(arg)
                    if isinstance(arg, dict):
                        for key in arg.keys():
                            for j in range(len(functions2[function])):
                                if isinstance(functions2[function][j], dict):
                                    if key not in functions2[function][j].keys():
                                        if function not in unique_keywords.keys():
                                            unique_keywords[function] = []
                                        unique_keywords[function].append(key)
                                    else:
                                        if arg[key] != functions2[function][j][key]:
                                            if function not in unique_keywords.keys:
                                                unique_keywords[function] = []
                                            unique_keywords[function].append(key)
    except SyntaxError:
        print("Syntax Error")
    return unique_function_calls, unique_params, unique_keywords
\end{lstlisting}
\end{figure*}

\section{Temperature Adjustments}
We present our experimentation results for the temperature tuning in Figure~\ref{fig:temp}.

\section{Example Dialogues}

We present the reference-image-based coder-director dialogues for the final example from Table \ref{tab:error_analysis}.

\begin{figure*}[ht]
\begin{tcolorbox}[width=\linewidth,colback={black!2!white},%
skin first=skin middle=skin]

\textbf{Coding Question:} highlight in red the x range 2 to 4 (0.667 Initial Pass@1)

\tcbline

\textbf{Cooperative Dialogue} (0.967 Final Pass@1)

\begin{addmargin}[1em]{0em}
\textbf{Director:}  highlight in red the x range 2 to 4

\textbf{Coder:}  Both solutions will highlight the x range from 2 to 4 in red, but they differ in transparency. Solution 1 uses an `alpha` of 0.3, which means it's more transparent than Solution 2 with an `alpha` of 0.5. If you want a less noticeable highlight, go with Solution 1; for more prominence, choose Solution 2.

\textbf{Director:} Increase the opacity of the highlighted area so that it's fully opaque instead of transparent. Make sure the highlight is prominently visible.

\textbf{Coder:}  To make the highlight fully opaque, you'll need to set the `alpha` parameter to 1. You can modify either of the existing solutions by changing `alpha=0.3` or `alpha=0.5` to `alpha=1`. This will eliminate transparency entirely and ensure that the red highlight is prominently visible.

\textbf{Director:} Make sure the highlight completely covers the x range from 2 to 4 with no gaps or transparency. The highlighted area should be clearly distinguished and fully filled in red.
\end{addmargin}

\tcbline

\textbf{Discoursive Dialogue} (1.0 Final Pass@1)

\begin{addmargin}[1em]{0em}
\textbf{Director:}  highlight in red the x range 2 to 4

\textbf{Coder:}  Clarification Question – The provided solutions differ in their transparency settings through the `alpha` values. Could you specify your preference regarding how prominent or subtle you'd like this red highlighting effect to be?

\textbf{Director:} Please ensure that the red highlight is fully opaque, with no transparency effect.

\textbf{Coder:}  Correction – To make sure we match your requirement of having no transparency, you'll need to modify the `alpha` value in either solution from its current setting to '1', achieving a fully opaque appearance.

\textbf{Director:} Please ensure the x range highlighted in red is fully opaque with no transparency. We need it to stand out distinctly against the rest of the plot.
\end{addmargin}

\tcbline

\textbf{Inquisitive Dialogue} (0.167 Final Pass@1)

\begin{addmargin}[1em]{0em}

\textbf{Director:}  highlight in red the x range 2 to 4

\textbf{Coder:}  What you seem to want is ensuring that the area between x=2 and x=4 is clearly visible on your plot highlighted in red. Do you want it to be very noticeable or somewhat subtle? The difference here is that one solution is more transparent than the other; which visibility level do you prefer?

\textbf{Director:} Please make the highlighted area more noticeable by removing any transparency so it's fully opaque.

\textbf{Coder:}  The implicit question here is about achieving complete opacity for the highlighted area between x=2 and x=4 on a plot. If neither existing solution (`Solution 1` or `Solution 2`) meets your requirements—since they include some degree of transparency—you'll need an alternative with no alpha setting or perhaps adjusting 'alpha' to `1`.

\textbf{Director:} Ensure the red highlighted area between x=2 and x=4 is fully opaque without any transparency. The `alpha` parameter should be set to 1, or simply not included at all to achieve full opacity.
    
\end{addmargin}

\end{tcolorbox}
\end{figure*}

\end{document}